\documentclass[runningheads]{llncs}

\usepackage{eccv}

\usepackage{eccvabbrv}

\usepackage{graphicx}
\usepackage{booktabs}
\usepackage{multirow}
\usepackage{placeins}

\usepackage[accsupp]{axessibility}  

\usepackage{hyperref}

\usepackage{orcidlink}

\begin{document}

\title{Estimating Velocity and Spin of Spherical Objects from Rolling-Shutter Image(s)} 

\titlerunning{Estimating Velocity of Spheres from Rolling-Shutter Image(s)}

\author{Wenjie Xue\inst{1}\orcidlink{0009-0008-9165-3539} \and
Jun Yang\inst{1}\orcidlink{0000-0001-5246-7671} \and
Jingmin Wang\inst{1,2} \and
Limin Shang\inst{1}}

\authorrunning{W. Xue et al.}

\institute{Epson Canada Ltd, Toronto, ON L3R 6G3, Canada \\
\email{\{Mark.Xue,Jun.Yang,Limin.Shang\}@ea.epson.com}\and
University of Toronto, Toronto, ON M5S 1A1, Canada \\
\email{jingmin.wang@mail.utoronto.ca}}
\renewcommand{\topfraction}{1}
\renewcommand{\bottomfraction}{0.8}
\renewcommand{\textfraction}{0.05}
\renewcommand{\floatpagefraction}{0.8}

\maketitle

\begin{abstract}
  Rolling-shutter cameras introduce characteristic distortions when imaging fast moving objects, and these effects are typically treated as artifacts to be corrected. In this work, we instead leverage rolling-shutter distortions as a valuable source of temporal information to estimate the 3D translational and angular velocities of rapidly moving spherical objects from a single rolling-shutter frame. We design a robust and easily detectable spherical pattern and propose a correspondence-free formulation that recovers motion by enforcing geometric consistency in a back-projection framework. By exploiting the geometry of the sphere, translational and rotational motions are decoupled and estimated through a two-stage optimization process, enabling reliable velocity recovery even for textureless objects. Extensive experiments on both synthetic and real datasets demonstrate accurate and robust estimation of motion parameters under challenging high-speed conditions.
\end{abstract}

\section{Introduction}
\label{sec:intro}

\begin{figure}[t]
  \centering
  \includegraphics[width=0.99\linewidth]{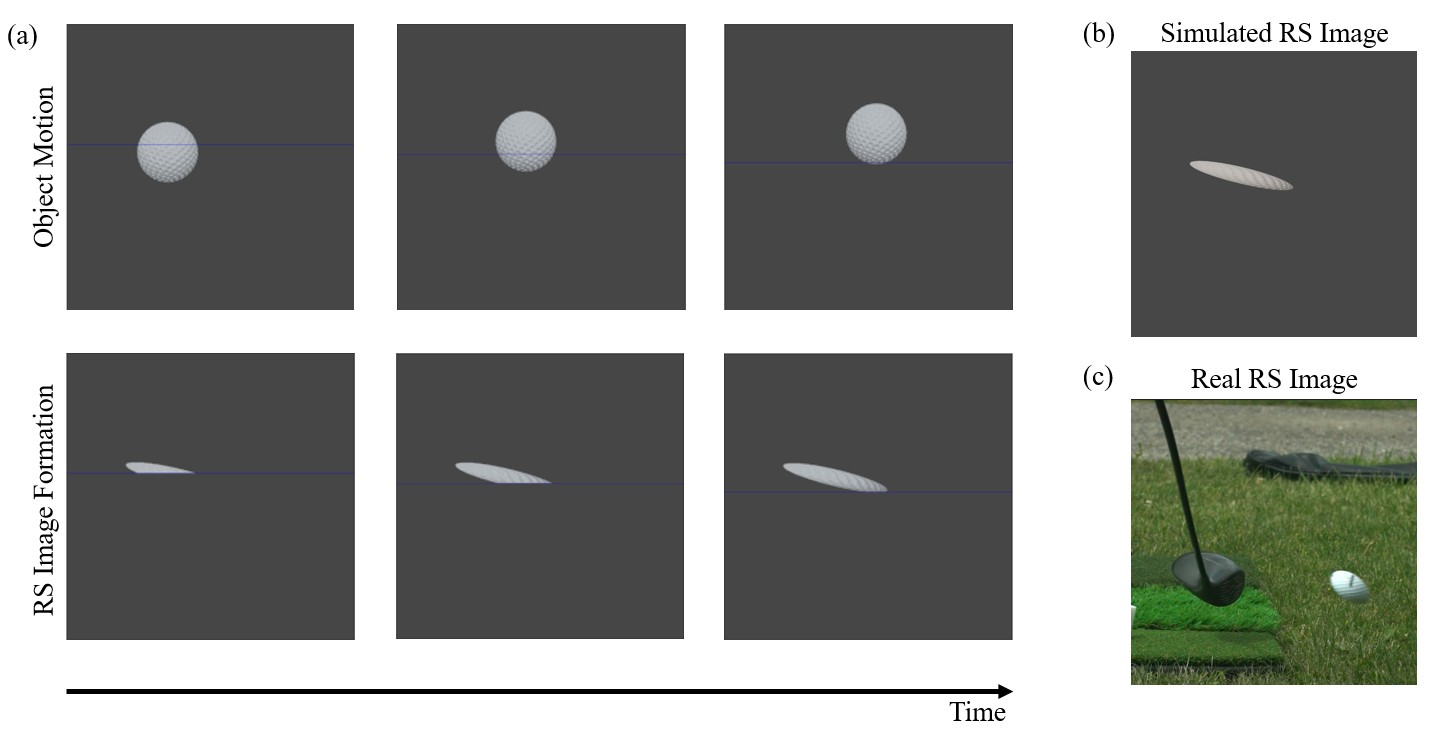}
  \vspace{-0.5\baselineskip}
  \caption{
    (a) Visualization of rolling-shutter distortion for a sphere moving rightward. Top: Scene motion during image acquisition. Bottom: Corresponding rolling-shutter image formation. The blue line indicates the active scanline. (b) Rolling-shutter image rendered by our simulator. (c) Rolling-shutter image captured in the real world.
  }
  \label{fig:intro}
\end{figure}

  Estimating the 3D translational velocity and spin of a fast-moving ball is central to sports and industrial applications. Currently, reliable motion capture typically relies on specialized high-speed global-shutter cameras, event sensors, or radar systems. By contrast, most commodity cameras use CMOS rolling-shutter sensors, whose sequential scanline exposure produces the familiar ``wobble'' artifacts under rapid motion (see \cref{fig:intro}). Existing rolling-shutter motion estimation methods, however, are a poor fit for spherical objects. Many approaches require explicit 3D--2D correspondences~\cite{Magerand2012RSObjectPoseMotion,AitAider2006RSPoseVelocity}, which are typically difficult to acquire in real-world scenarios. Other methods rely on straight-line features with additional parameters and assumptions~\cite{AitAider2007KinematicsFromLines}, limiting their applicability to spheres, which often lack such straight-line structures. Moreover, these formulations jointly optimize a large number of coupled variables, leading to highly non-convex problems that are sensitive to initialization and prone to instability.

  In this work, we present a formulation that eliminates the need for explicit 2D--3D correspondences and leverages rolling-shutter distortions as motion cues for spherical objects. Our key idea is to back-project observed image features into 3D rays, transform them to a common temporal reference frame, and recover object motion through geometric consistency constraints. To achieve this, we first design a robust, easily detectable pattern for the sphere and formulate 6D sphere motion estimation as a two-stage optimization problem. In the first stage, we estimate the 3D translational motion by exploiting the sphere's geometric properties, making the method applicable to textureless objects. In the second stage, we leverage the designed pattern to estimate the sphere's 3D rotational motion through a separate optimization. The resulting optimization is low-dimensional, computationally efficient, and stable in practice, while naturally extending to multi-image and multi-camera settings. Together, these design choices yield a robust and efficient solution for spherical motion recovery.

  In summary, we make the following key contributions:
  \begin{enumerate}
    \item A back-projection-based algorithm that estimates 3D translational and angular velocity from a single rolling-shutter image without requiring explicit 3D–2D correspondences.
    \item A geometry-driven two-stage optimization framework that decouples translational and angular motion, yielding stable convergence and reliable translation estimates for textureless spheres.
    \item Extensions to multi-image and multi-camera configurations, validated in both simulation and a real-world golf launch-monitor prototype.
  \end{enumerate}

\section{Related Work}
\subsection{Rolling-Shutter Cameras}

Rolling-shutter cameras capture image rows sequentially, producing geometric distortions in dynamic scenes~\cite{Meingast2005GeometricModelsRollingShutter}. Most prior work treats these distortions as artifacts to be removed~\cite{Baker2010RemovingRollingShutterWobble,Grundmann2012CalibrationFreeRSRemoval}. Other approaches explicitly model rolling-shutter geometry within pose estimation and bundle adjustment frameworks~\cite{Hedborg2012RSBA,liao2023revisitingrollingshutterbundle,Dai2016RSCameraRelativePose,Saurer2015MinimalRSPose,Albl2015R6PRollingShutterAbsolutePose,Albl2020RollingShutterCameraAbsolutePose}. More recently, learning-based methods have been proposed to estimate camera motion and synthesize undistorted images from one or more rolling-shutter frames~\cite{Rengarajan2017UnrollingtheShutter,Liu2020DeepShutterUnrolling}.

Beyond correction, a complementary line of research exploits rolling-shutter distortions as motion cues. Ait-Aider et al.~\cite{AitAider2006RSPoseVelocity} showed that pose and velocity can be recovered from a single RS image with 2D–3D correspondences. Later works extended this idea to more complex motion models, including piecewise global-shutter approximations for non-uniform motion~\cite{Magerand2010AGR}, Taylor expansions for small angular motion~\cite{Magerand2012RSObjectPoseMotion}, and geometric primitives such as line correspondences for kinematic estimation~\cite{AitAider2007KinematicsFromLines}.

Despite these advances, the above methods rely on accurate feature correspondences, which are difficult to obtain for fast-moving spherical objects that are often textureless or exhibit repetitive patterns. Moreover, in industrial and sports settings, rapid object motion violates the small-motion and global-shutter assumptions of prior methods, degrading feature detection and matching. As a result, no dedicated formulation robustly recovers spherical object motion directly from rolling-shutter imagery.

\subsection{Motion Estimation for Spherical Objects}
Estimating the velocity of a moving sphere has been widely studied in computer vision and robotics, typically using multi-frame or multi-camera setups with high-speed global-shutter cameras to avoid rolling-shutter distortions~\cite{Tamaki2004BallSpinRegistration,Tamaki2012TableTennis,Cant2020Tennis}. In parallel, event-camera-based approaches have emerged as an alternative that bypasses these limitations, enabling accurate recovery of spherical motion through spatiotemporal optimization of asynchronous events~\cite{Nakabayashi2024EventCamera}. More recently, learning-based methods have leveraged video sequences and known scene geometry to infer 3D trajectory and spin directly from data~\cite{kienzle2025ballspintrajectoryanalysis}. Visual servoing–based approaches~\cite{chaumette2006visual} have also been explored in robotic settings, where image feedback is used to estimate and track sphere motion online in a closed-loop manner~\cite{dahmouche2012dynamic}.

In contrast, our work does not require costly robotic setups and demonstrates that sphere velocity can be recovered using a low-cost rolling-shutter camera.

\section{Problem Formulation}
  For a fast-moving spherical object, we aim to estimate its translational velocity, $\mathbf{v}_{co}\in\mathbb{R}^{3}$, and angular velocity, $\boldsymbol{\omega}_{co}\in\mathbb{R}^{3}$, from measurements captured by a rolling-shutter camera. For global-shutter cameras, which capture the entire frame instantaneously, the estimation of these velocities can be formulated as a simple nonlinear optimization problem using 2D–3D correspondences.

  Specifically, for a 3D object point, $\mathbf{P}_o\in\mathbb{R}^{3}$, defined in the object frame $\mathcal{F}_o$, its position in the camera frame, $\mathcal{F}_c$, at timestamp $k$ is defined as:
  \begin{equation}
  \mathbf{P}_{c,k}=\mathbf{R}_{co,k}\mathbf{P}_o+\mathbf{t}_{co, k},
  \end{equation}  
  where $\mathbf{R}_{co,k}\in\mathbb{SO}(3)$ and $\mathbf{t}_{co, k}\in\mathbb{R}^{3}$ denote the rotation and translation of the object with respect to the camera frame at timestamp k, respectively. For brevity, we will omit the subscript “$co$” in the remainder of this section. Due to the short exposure time relative to the high-speed sphere motion, we assume constant velocity over a short time interval $\Delta t_g$ from timestamp $0$ to $k$. Under this assumption, the pose $\left(\mathbf{R}_k, \mathbf{t}_k\right)$ evolve as follows:
  \begin{equation}
  \label{equ:rotation}
  \mathbf{R}_k = \exp\!\left(\boldsymbol{\omega}_{co}^{\wedge}\Delta t_g\right)\mathbf{R}_0 ,
  \end{equation}
  \vspace{-1.0\baselineskip}
  \begin{equation}
  \label{equ:translation}
  \mathbf{t}_k = \mathbf{t}_0 + \mathbf{v}_{co}\Delta t_g,
  \end{equation}  
  where $\mathbf{R}_0$ and $\mathbf{t}_0$ denote the rotation and translation of the sphere at timestamp $0$. The operator $\exp(\cdot)$ denotes the exponential map from the Lie algebra $\mathfrak{so}(3)$ to the Lie group $\mathbb{SO}(3)$, and ${(\cdot)}^\wedge$ is the skew-symmetric operator that converts a vector in $\mathbb{R}^3$ into a matrix in $\mathfrak{so}(3)$. Given the corresponding 2D image measurement, $\mathbf{u}={\left[u_x,u_y\right]}^\top$, of the 3D point $\mathbf{P}_o$, we can construct the following nonlinear optimization problem to estimate the velocities:
  \begin{equation}
  \label{equ:global_optimization}
  \mathbf{v}_{co}, \boldsymbol{\omega}_{co} = \arg\min_{\mathbf{v}_{co}, \boldsymbol{\omega}_{co}} 
  \sum_{i=0}^{N-1} 
  \left\| 
  \pi\big( \mathbf{R}_k \mathbf{P}_o^i + \mathbf{t}_k \big) - \mathbf{u}^i 
  \right\|^2,
  \end{equation}  
  with \(i \in [0, N-1]\) indexing the 3D points on the object, and $\pi(\cdot)$ denotes the pinhole camera projection of a 3D point using the camera intrinsic matrix $\mathbf{K}$. This optimization, for a global-shutter camera, minimizes the reprojection error to estimate the sphere’s translational velocity $\mathbf{v}_{co}$ and angular velocity $\boldsymbol{\omega}_{co}$, assuming all pixels are captured simultaneously.

  In contrast, a rolling-shutter camera acquires image rows sequentially, meaning that different rows correspond to different time instants. As a result, the image contains motion-induced distortion that must be modeled to accurately estimate $\mathbf{v}_{co}$ and $\boldsymbol{\omega}_{co}$. To this end, we introduce the parameter $\tau$, representing the readout time between consecutive scanlines, which can be calibrated using a rapidly flashing LED. Unlike a global shutter, where the entire image is captured after a fixed interval $\Delta t_g$, the effective capture time of a specific row $\Delta t_r\left(\mathbf{u}\right)$ in a rolling-shutter image becomes:
  \begin{align}
  \label{equ:scanline}
  \Delta t_r\left(\mathbf{u}\right) &= \Delta t_g + \tau \left(u_y - u_y(0)\right)\\
  &=\Delta t_g + \tau u_y,
  \end{align}
  where $u_y - u_y(0)$ denotes the row offset from the first scanline, i.e., the vertical pixel distance from the first image row to the current row. Substituting \cref{equ:scanline} into \cref{equ:rotation} and \cref{equ:translation} extends the optimization in \cref{equ:global_optimization} to account for rolling-shutter effects. 
  
  However, solving such optimization in practice is challenging because the capture time $\Delta t(\mathbf{u})$ depends on the unknown pixel coordinate $u_y$, which in turn depends on the 3D-to-2D projection. This coupling creates an implicit nonlinear problem for each feature point, making direct optimization computationally difficult. To address this, we employ a backward-projection formulation, as detailed in the next section.

\section{Methodology}
\begin{figure}[t]
  \centering
  \includegraphics[width=0.9\linewidth]{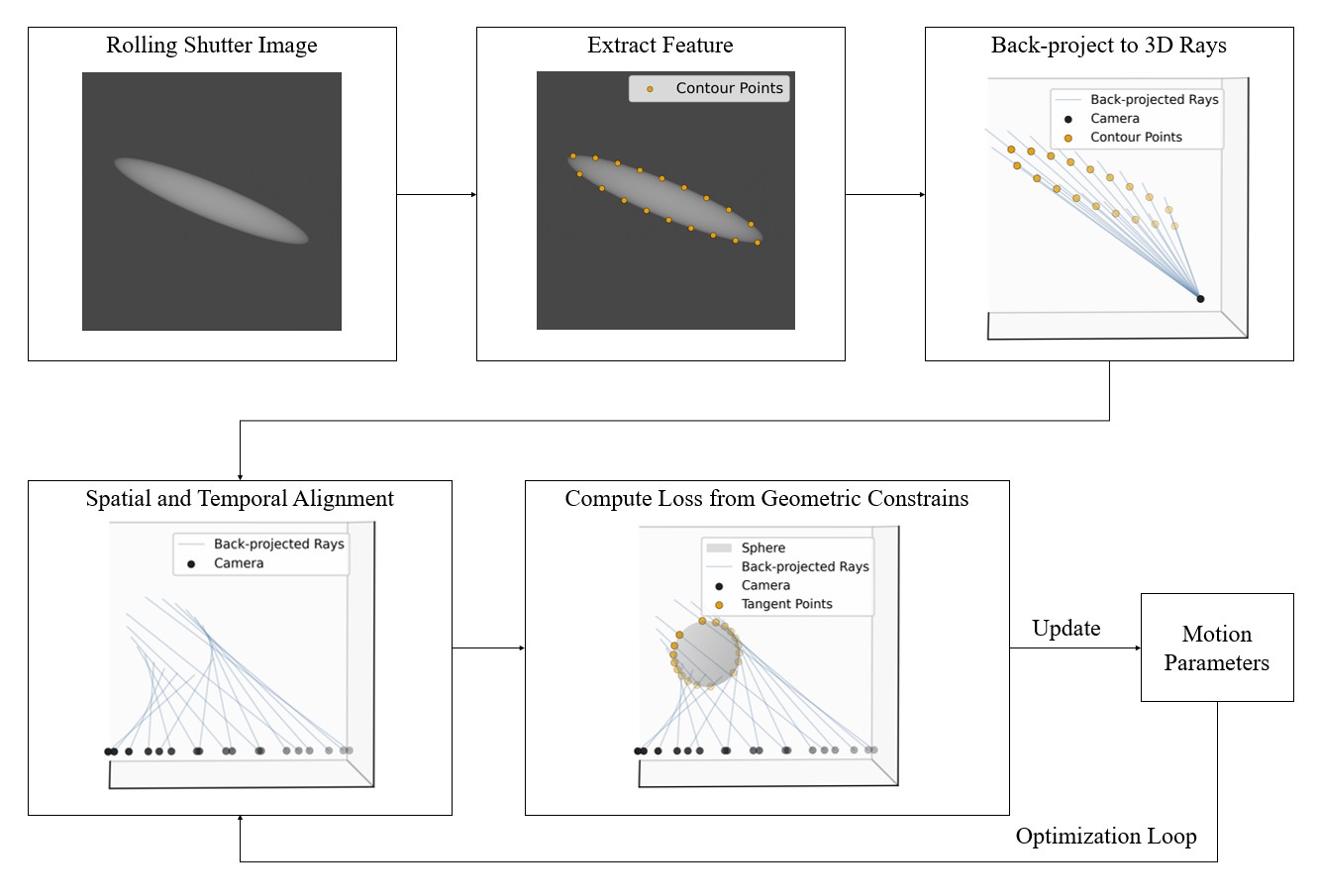}
  \vspace{-0.85\baselineskip}
  \caption{
    Illustration of our pipeline for estimating the sphere’s translational velocity. Detected image keypoints are first back-projected to 3D rays in object coordinates. The rays are then temporally and spatially aligned under the current motion parameters, and a loss is computed from the aligned rays to update the parameters. Spin estimation follows the same procedure.
  }
  \label{fig:pipeline}
\end{figure}

In this section, we present our approach for estimating spherical motion under rolling-shutter image formation. To account for row-dependent capture timing, we adopt a back-projection formulation and decompose the problem into two decoupled optimizations for translational and angular velocities. An overview of the proposed pipeline is shown in \cref{fig:pipeline}. In the following sections, we first describe the back-projection formulation and coordinate representation in \cref{sec:back_proj}. We then detail the estimation of translational and angular velocities in \cref{sec:trans} and \cref{sec:rot}, respectively. Finally, we extend our framework to the multi-camera and multi-frame setting in \cref{sec:multi_view}.

\subsection{Back-Projection Formulation} 
\label{sec:back_proj}
By the principle of relative motion, observing a moving sphere with a static camera is equivalent to observing a static sphere with a moving camera. For consistency in our formulation and to simplify subsequent motion estimation, we express all quantities in the object coordinate frame $\mathcal{F}_o$.

For a 2D image measurement, $\mathbf{u}\in\mathbb{R}^{2}$, captured by a rolling-shutter camera, the pixel defines a back-projection ray originating from the camera optical center $\mathbf{x}\in\mathbb{R}^{3}$ and passing through the image plane. Expressed in the object frame $\mathcal{F}_o$, the ray is parameterized as:
\begin{equation} \label{equ:back_proj_begin}
\mathbf{r}_o(a,\mathbf{u}) = \mathbf{x}_o + a\,\mathbf{d}_o(\mathbf{u}), \quad a > 0,
\end{equation}
where $a$ denotes the depth along the ray, $\mathbf{x}_o\in\mathbb{R}^{3}$ is the camera center, and $\mathbf{d}_o\in\mathbb{R}^{3}$ is the back-projected viewing direction, both expressed in the object frame. The camera center and viewing direction are given by:
\begin{align} 
\mathbf{x}_o &= \mathbf{t}_{oc} = -\mathbf{R}_{co}^\top\mathbf{t}_{co} \label{equ:origin} \\
\mathbf{d}_o(\mathbf{u}) &\propto \mathbf{R}_{oc} \mathbf{K}^{-1} \begin{bmatrix} u_x \\ u_y \\ 1 \end{bmatrix}\:\:,\:\:\lVert\mathbf{d}_o(\mathbf{u})\rVert=1 \label{equ:direction}
\end{align}
where $\mathbf{K}$ denotes the camera intrinsic matrix, and $\mathbf{t}_{oc}$ and $\mathbf{R}_{oc}$ represent the translation and rotation from the camera frame to the object frame, respectively.

For a moving camera that starts motion at a reference timestamp $k_0$, the elapsed time for a pixel measurement $\mathbf{u}={\left[u_x,u_y\right]}^\top$ captured at timestamp $k$ by a rolling-shutter camera is:
\begin{align}
\Delta t\left(\mathbf{u}\right) &= k- k_0 \\
\label{equ:delta_t}
&= \left(k_s +\tau u_y \right)- k_0 ,
\end{align}
where $k_s$ is the timestamp of the first scanline and $\tau$ is the readout time per row. Due to the rolling-shutter effect, each image row has a slightly different capture time, distorting rigid shapes. To preserve the geometry under rigid-body motion, we align all rays from time $k$ to the reference timestamp $k_0$. 

Let $\mathbf{v}_{oc}$ and $\boldsymbol{\omega}_{oc}$ denote the camera's translational and rotational velocities, related to the object's motion by $\mathbf{v}_{oc} = -\mathbf{R}_{co}^\top\mathbf{v}_{co}$ and $\boldsymbol{\omega}_{oc} = -\mathbf{R}_{co}^\top\boldsymbol{\omega}_{co}$. The aligned ray position and direction are then given by:
\begin{align}
\label{equ:ray1}
    \mathbf{x}_{o, k_0}(\mathbf{u}, \mathbf{t}_{co}, \mathbf{v}_{co}, \boldsymbol{\omega}_{co}) &=\exp\!\left(\Delta t(\mathbf{u})\boldsymbol{\omega}_{oc}^{\wedge}\right)\left(\mathbf{x}_{o}+\Delta t(\mathbf{u})\mathbf{v}_{oc}\right) \\
    &=\exp\!\left(-\Delta t(\mathbf{u})(\mathbf{R}_{co}^\top\boldsymbol{\omega}_{co})^\wedge\right)\left(\mathbf{x}_{o}-\Delta t(\mathbf{u})\mathbf{R}_{co}^\top\mathbf{v}_{co}\right). \\
    \label{equ:ray2}
    \mathbf{d}_{o,k_0}(\mathbf{u}, \mathbf{t}_{co}, \mathbf{v}_{co}, \boldsymbol{\omega}_{co}) &= \exp\!\left(\Delta t(\mathbf{u})\boldsymbol{\omega}_{oc}^{\wedge}\right)\mathbf{d}_o(\mathbf{u}) \\
    &= \exp\!\left(-\Delta t(\mathbf{u})(\mathbf{R}_{co}^\top\boldsymbol{\omega}_{co})^{\wedge}\right)\mathbf{d}_o(\mathbf{u}), 
\end{align}

For notational clarity, we omit the motion parameters $\mathbf{t}_{co}, \mathbf{v}_{co}, \boldsymbol{\omega}_{co}$ in the remainder of this paper. For pixel measurement $\mathbf{u}={\left[u_x,u_y\right]}^\top$, the back-projection ray at the reference timestamp $k_0$ is finally represented as:
\begin{equation} 
\label{equ:final_ray}
\mathbf{r}_{o,k_0}\left(a;\mathbf{u}\right) = \mathbf{x}_{o,k_0}\left(\mathbf{u}\right) + a \mathbf{d}_{o,k_0}\left(\mathbf{u}\right), \quad a > 0, 
\end{equation}
This backward projection formulation enables the estimation of both translational and rotational motion without requiring explicit 3D–2D correspondences~\cite{AitAider2006RSPoseVelocity,Magerand2012RSObjectPoseMotion}, as will be described in \cref{sec:trans,sec:rot}.

\subsection{Translational Velocity Estimation} 
\label{sec:trans}
To estimate the translational velocity, as illustrated in \cref{fig:pipeline}, we extract 2D contour points of the observed sphere. Since the sphere's silhouette is invariant to rotation, the resulting contour constraints depend only on the sphere's position $\mathbf{t}_{co}$ and translational velocity $\mathbf{v}_{co}$. This enables a two-stage estimation procedure: we first determine the object position and translational velocity, then recover the angular velocity. This decomposition reduces the number of jointly optimized parameters from 12 to 6, improving optimization efficiency.

As illustrated in \cref{fig:pipeline}, let $\mathbf{z} = [\mathbf{u}_{0}^\top, \dots, \mathbf{u}_{N-1}^\top]^\top$ be the set of detected 2D contour points, where $\mathbf{u}_{i}$ denotes the $i$-th measured point along the contour. For each contour measurement, we construct the projection ray using \cref{equ:back_proj_begin}–\cref{equ:final_ray} under a motion hypothesis with zero angular velocity, yielding:
\begin{equation}
\mathbf{r}_{o,k_0}\left(a; \mathbf{u}_i\right) = \mathbf{x}_{o,k_0}\left(\mathbf{u}_i\right) + a \mathbf{d}_{o,k_0}\left(\mathbf{u}_i\right), \quad a > 0, 
\end{equation}
where $\mathbf{r}_{o,k_0}\left(a;\mathbf{u}_i\right)$ represents the ray associated with the $i$-th contour point. For brevity, we will omit the subscript “$o,k_0$” and use subscript “$i$” to denote correspondence to $\mathbf{u}_i$ in the following sections.

Along this ray, the 3D point that minimizes the Euclidean distance to the sphere center can be computed as:
\begin{equation}
  \hat{a}_i = \arg\min_{a\in\mathbb{R}^+}\left\lVert \mathbf{x}_i + a\,\mathbf{d}_i \right\rVert,
\end{equation}
which has the closed-form solution:
\begin{equation}
  \hat{a}_i = \max \left(0, - \mathbf{d}_i^{\top} \mathbf{x}_i\right) \:,\: \mathbf{P}_i=\mathbf{x}_i + \hat{a}_i\,\mathbf{d}_i.
\end{equation}
where $\mathbf{P}_i$ is the obtained 3D point on the ray.

In the ideal case, when the sphere's initial position $\mathbf{t}_{co}$ and translational velocity, $\mathbf{v}_{co}$ are correct, the resulting 3D point $\mathbf{P}_i$ from the previous equation lies on the sphere surface, satisfying:
\begin{equation}
  \lVert\mathbf{P}_i\rVert = r,
\end{equation}
where $r$ is the known sphere radius. To enforce this, we define the velocity-consistency loss:
\begin{equation}
  \mathcal{L}_\mathbf{v}(\mathbf{t}_{co}, \mathbf{v}_{co}) = \frac{1}{N} \sum_{i=0}^{N-1} \big( \lVert \mathbf{P}_i \rVert - r \big)^2.
\end{equation}
However, relying solely on this velocity loss may lead to a degenerate solution, where all rays collapse into a small region on the sphere surface. To prevent this, we introduce a regularization term that enforces a sufficient spatial spread of the reconstructed points:
\begin{equation}
\mathcal{R}_{\mathbf{v}}(\mathbf{t}_{co}, \mathbf{v}_{co}) =
\max\Bigg(
0,\;
\sigma^{2} r^{2}
-
\frac{1}{N} \sum_{i=0}^{N-1} 
\left\lVert \mathbf{P}_i - \bar{\mathbf{P}} \right\rVert^{2}
\Bigg),
\end{equation}
where $\bar{\mathbf{P}}$ is the mean of the reconstructed 3D points, and $\sigma$ determines the minimum allowable spread of the points, set empirically to $\sigma = \frac{1}{2}$. Our final objective is to minimize the following loss using the L-BFGS-B algorithm~\cite{Byrd1995LBFGSB}:
\begin{equation} \label{eq:velocity_loss}
  \mathcal{J}_\mathbf{v}(\mathbf{t}_{co}, \mathbf{v}_{co}) =
  \mathcal{L}_\mathbf{v}(\mathbf{t}_{co}, \mathbf{v}_{co}) +
  \mathcal{R}_\mathbf{v}(\mathbf{t}_{co}, \mathbf{v}_{co}).
\end{equation}

\begin{figure}[t]
  \centering
  \includegraphics[width=0.24\linewidth]{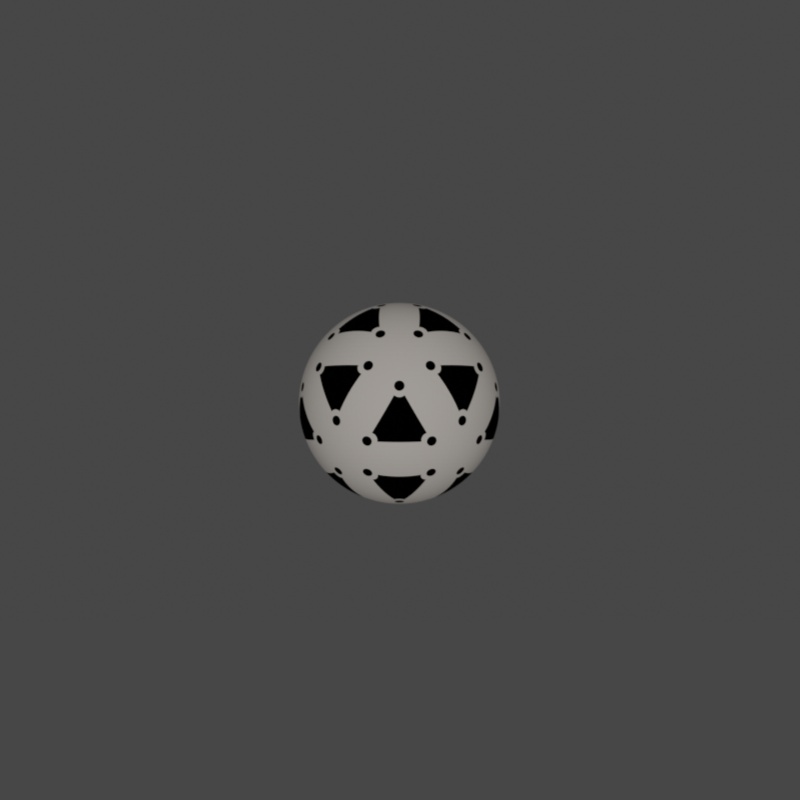}\hfill
  \includegraphics[width=0.24\linewidth]{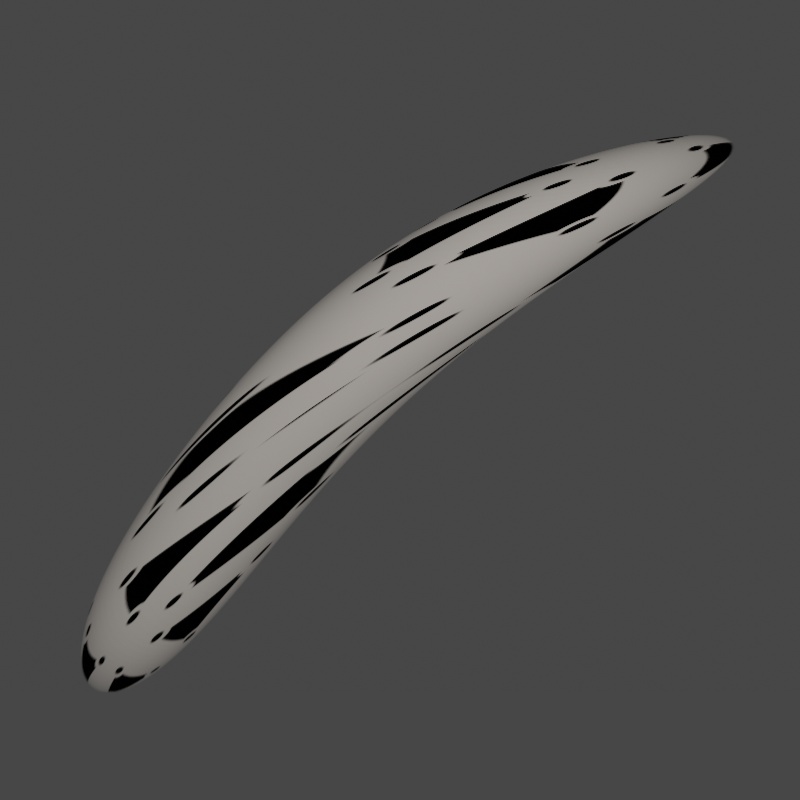}\hfill
  \includegraphics[width=0.24\linewidth]{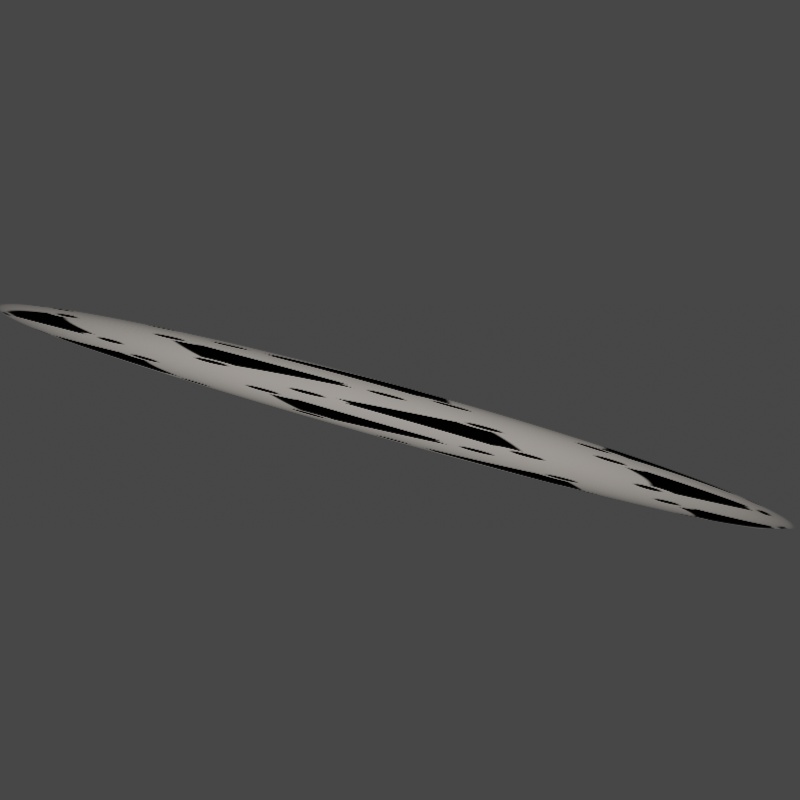}\hfill
  \includegraphics[width=0.24\linewidth]{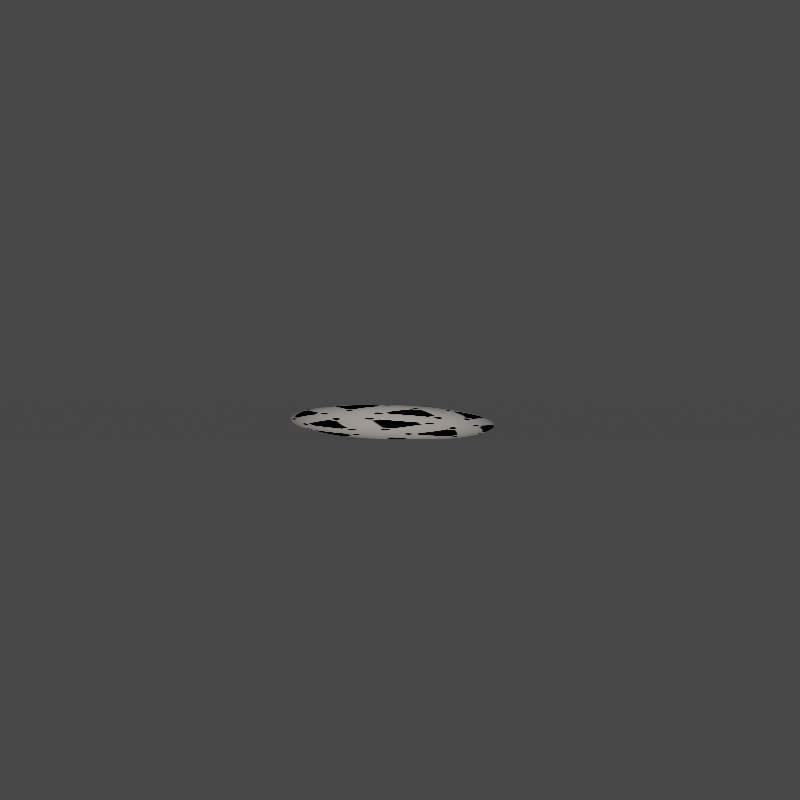}
  \vspace{-0.5\baselineskip}
  \caption{Our surface pattern design for the sphere. The leftmost image shows the stationary sphere, while the remaining three are representative rolling-shutter images.}
  \label{fig:sample_data}
\end{figure}

\subsection{Angular Velocity Estimation} 
\label{sec:rot}
To recover the angular velocity $\boldsymbol{\omega}_{co}$, we design a structured surface pattern on the sphere (\cref{fig:sample_data}) and estimate it by enforcing relative surface feature constraints. Specifically, we assume a set of salient surface features arranged into point pairs with equal distances. Under the rolling-shutter motion model, these pairwise distances are constrained to remain constant, enabling the recovery of $\boldsymbol{\omega}_{co}$.

Let $z=[\mathbf{u}_0^\top, ..., \mathbf{u}_{N-1}^\top]^\top$ denote the set of detected surface feature points and $\mathcal{E}\subset \{0,...,N-1\}^2$ denote the set of equidistant point pairs. Similar to \cref{sec:trans}, we back-project feature point $\mathbf{u}_i$ to 3D ray $\mathbf{r}_i$ under a motion hypothesis in which $\mathbf{t}_{co}$ and $\mathbf{v}_{co}$ are determined from the previous step. The first intersection between $\mathbf{r}_i$ and the sphere of radius $r$ is obtained by solving 
\begin{equation}
  \left\lVert \mathbf{x}_i + \hat{a}_i\,\mathbf{d}_i \right\rVert = r,
\end{equation}
which yields the following closed-form expression for the smaller root
\begin{equation}
  \hat{a}_i = -\mathbf{d}_i^{\top}\mathbf{x}_i - \sqrt{\left(\mathbf{d}_i^{\top}\mathbf{x}_i\right)^2 - \left(\left\lVert\mathbf{x}_i\right\rVert^2 - r^2\right)}\:\:,\:\:\mathbf{P}_i=\mathbf{x}_i+\hat{a}_i\mathbf{d}_i,
\end{equation}
where $\mathbf{P}_i$ is the obtained surface point.
We then construct the objective as the variance of the point-pair distances:
\begin{equation}
\mathcal{J}_\omega(\boldsymbol{\omega}_{co})
=
\frac{1}{|\mathcal{E}|}
\sum_{(i,j)\in\mathcal{E}}
\left\lVert \mathbf{P}_i - \mathbf{P}_j \right\rVert^2
-
\left(
\frac{1}{|\mathcal{E}|}
\sum_{(i,j)\in\mathcal{E}}
\left\lVert \mathbf{P}_i - \mathbf{P}_j \right\rVert
\right)^2.
\end{equation}

\subsection{Extension to Multi-Camera and Multi-Frame} 
\label{sec:multi_view}

Our formulation extends straightforwardly to multiple images, either captured by a single or multiple rolling-shutter cameras. Let $\{I_{c_0},...,I_{c_{N-1}}\}$ be the images captured by cameras ${c_0, ..., c_{N-1}}$. The relative rotation, $R_{c_0c_i}$, and relative translation, $\mathbf{t}_{c_0c_i}$, of camera $c_i$ with respect to $c_0$ can be obtained through calibration. To align back-projected rays to object frame $\mathcal{F}_o$, we modify \cref{equ:origin,equ:direction} to 
\begin{align} 
\mathbf{x}_o &= \mathbf{R}_{c_0o}^\top(\mathbf{t}_{c_0c_i} - \mathbf{t}_{c_0o}) \\
\mathbf{d}_o(\mathbf{u}) &\propto \mathbf{R}_{c_0o}^\top \mathbf{R}_{c_0c_i} \mathbf{K}^{-1}_{c_i} \begin{bmatrix} u_x \\ u_y \\ 1 \end{bmatrix}\:\:,\:\:\lVert\mathbf{d}_o(\mathbf{u})\rVert=1
\end{align}
where $\mathbf{K}_{c_i}$ is the intrinsic matrix of camera $c_i$. The subsequent derivation of back-projection follows identically to the single-camera case.

To solve for translational velocity under a multi-image scenario, we calculate objective $\mathcal{J}_\mathbf{v}^{(i)}(\mathbf{t}_{c_0o},\mathbf{v}_{c_0o})$ for each image $I_i$ and jointly optimize the objective as:
\begin{equation}
  \mathcal{J}_\mathbf{v}(\mathbf{t}_{c_0o},\mathbf{v}_{c_0o}) = \sum_{i=0}^{N-1}\mathcal{J}_\mathbf{v}^{(i)}(\mathbf{t}_{c_0o},\mathbf{v}_{c_0o}).
\end{equation}

Based on the previously estimated motion parameters, we subsequently optimize the angular velocity in a similar manner using the L-BFGS-B solver~\cite{Byrd1995LBFGSB}:
\begin{equation} \label{equ:multi_angular_loss}
  \mathcal{J}_{\boldsymbol{\omega}, 1}(\boldsymbol{\omega}_{c_0o}) = \sum_{i=0}^{N-1}\mathcal{J}_{\boldsymbol{\omega}}^{(i)}(\boldsymbol{\omega}_{c_0o}).
\end{equation}

When the images are captured with an inter-frame delay, we further refine the estimate by exploiting the larger temporal baseline. Let $\mathcal{P}_i$ denote the projected surface points corresponding to image $I_i$. We define an inter-frame objective based on the truncated Chamfer distance. The one-way truncated squared Chamfer distance from the set $\mathcal{A}$ to the set $\mathcal{B}$ is given by
\begin{equation}
d_{\mathrm{ch}}(\mathcal{A},\mathcal{B})
=
\frac{1}{|\mathcal{A}|}
\sum_{\boldsymbol{a}\in\mathcal{A}}
\min\!\left(
D,\;
\min_{\boldsymbol{b}\in\mathcal{B}}
\left\lVert \boldsymbol{a}-\boldsymbol{b} \right\rVert
\right)^2,
\end{equation}
where $D$ is a truncation hyper-parameter that limits the pairwise matching distance to mitigate the effect of incomplete correspondences between frames. Using this metric, we introduce the refinement loss:

\begin{equation}
\mathcal{J}_{\boldsymbol{\omega},2}(\boldsymbol{\omega}_{c_0o})
=
\sum_{i=0}^{N-1}\sum_{j=0}^{N-1}
d_{\mathrm{ch}}\!\left(\mathcal{P}_i,\mathcal{P}_j\right).
\end{equation}

\begin{figure}[t]
  \centering
  \includegraphics[width=0.8\linewidth]{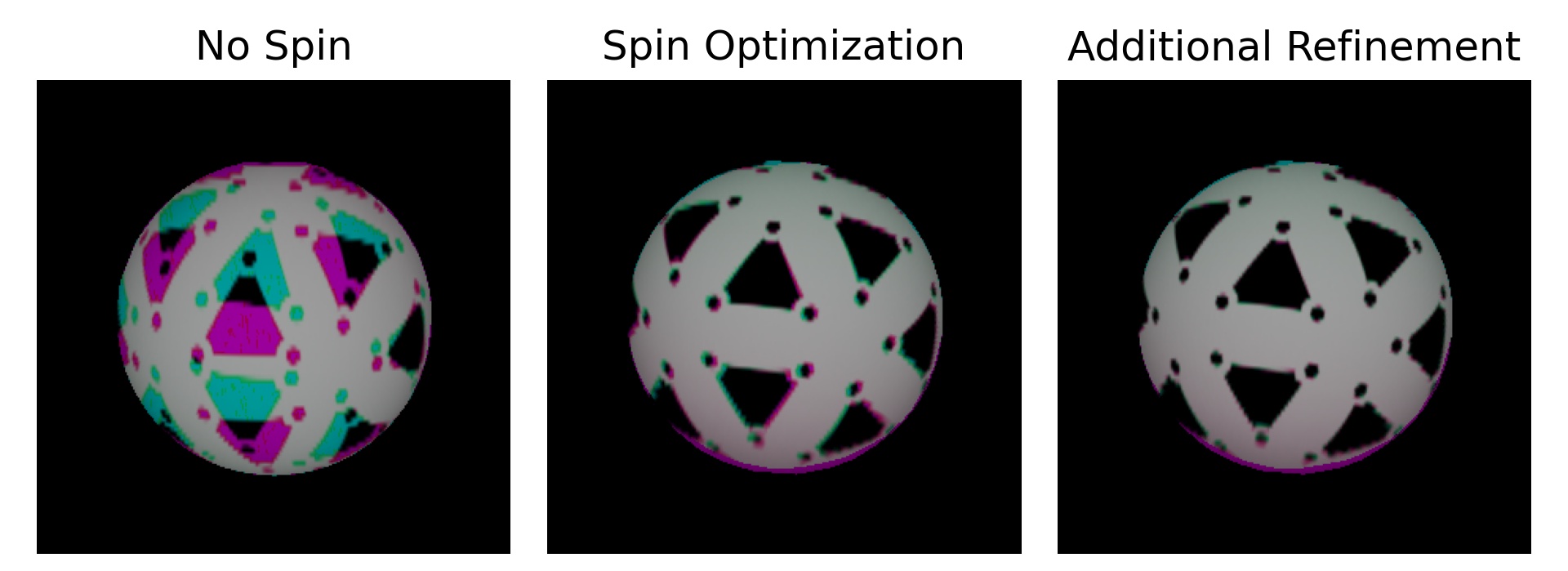}  
  \vspace{-0.7\baselineskip}
  \caption{Proposed spin refinement step. Left: overlay of two frames captured 2\,ms apart. Middle: alignment after unwarping using spin estimated from geometric constraints only, with minor residual misalignment remaining. Right: alignment after the refinement step, resulting in near-perfect pattern overlap.}
  \label{fig:multi_spin}
\end{figure}

The result of \cref{equ:multi_angular_loss} serves as the initialization for this refinement step. The objective $\mathcal{J}_{\boldsymbol{\omega},2}$ enforces temporal consistency by aligning the back-projected surface points across multiple frames. We illustrate this effect in \cref{fig:multi_spin}.

\section{Experimental Setup}
We describe the experimental setup used for evaluating the proposed method, first in simulation and then in real-world scenarios.
\subsection{Pattern Design} 
\label{sec:pattern}

\begin{figure}[t]
  \centering
  \includegraphics[width=\linewidth]{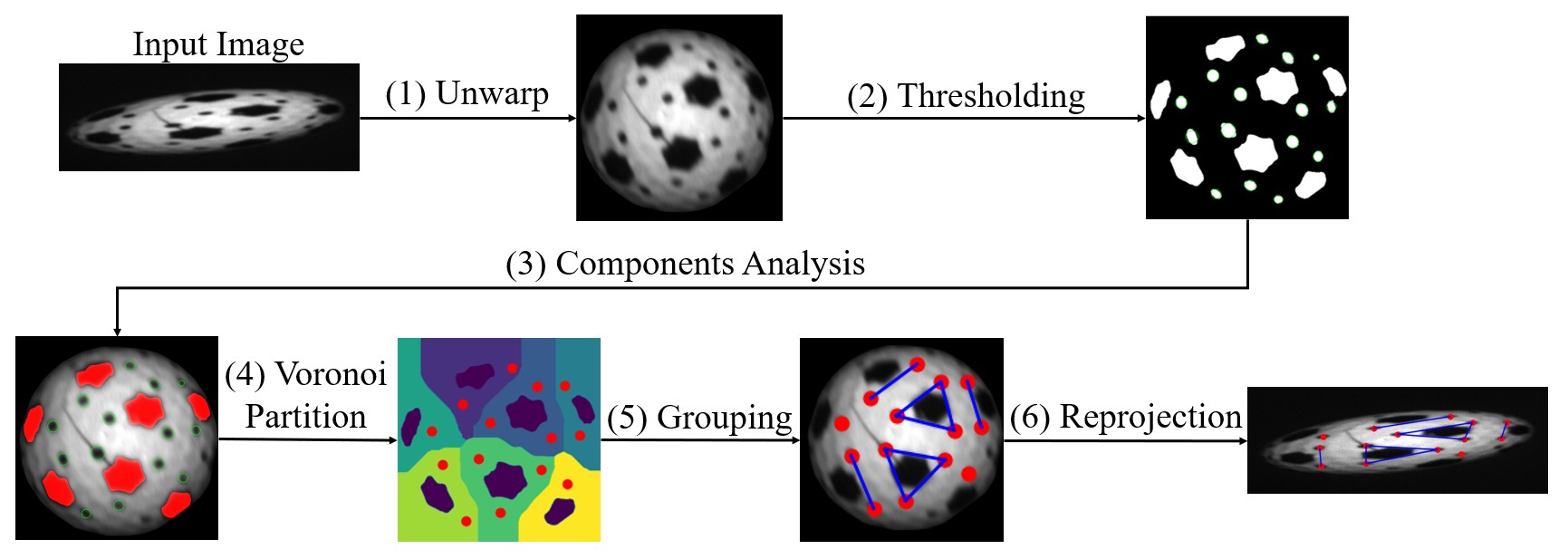}
  \vspace{-1.5\baselineskip}  
  \caption{
    Feature-extraction pipeline for the designed pattern. (1) Unwrap the RS image using the predicted velocity. (2) Apply adaptive thresholding to segment the pattern. (3) Classify connected components as dots or markers. (4) Partition the image into Voronoi cells induced by the markers. (5) Group dots into edges according to the Voronoi cells. (6) Reproject the detected feature coordinates into the original image.
  }
  \label{fig:feature_extraction}
\end{figure}

As discussed in~\cref{sec:intro} and \cref{sec:rot}, recovering rotational velocity requires surface features on the sphere. We therefore design a robust, easily detectable surface pattern, illustrated in \cref{fig:sample_data}. The pattern is derived from an icosahedron and consists of 20 groups of dots, each forming an equilateral triangle with a shaded interior for robust grouping. The shaded triangles provide a robust cue for grouping the dots, enabling reliable detection of equidistant point pairs by the feature extraction pipeline shown in \cref{fig:feature_extraction}.

The icosahedron-inspired layout ensures that multiple triangles remain fully visible from most viewpoints, providing stable geometric constraints for spin estimation. In rare edge cases, a triangle vertex may lie near the silhouette and become difficult to detect; therefore, we slightly shrink the triangles to increase the number of reliably visible points.

\subsection{Rolling-Shutter Simulation and Baseline} \label{sec:simulation}

We implement a rolling-shutter simulator in Blender~\cite{Blender}. For each trial, a patterned sphere is initialized with a random pose, and its motion is simulated under constant translational and angular velocities. A sequence of intermediate global-shutter frames is rendered, from which a single rolling-shutter image is synthesized by selecting the corresponding scanlines. Representative simulated rolling-shutter images are shown in \cref{fig:sample_data}. In our simulation, we set the camera resolution to $2048 \times 2048$, the focal length to 3000 pixels, and the readout time to $10\,\mu$s. The camera is positioned 30\,cm from the sphere. We sample translational and angular velocity directions uniformly on the unit sphere, with magnitudes drawn uniformly from 0--50\,m/s and 0--5000\,rpm, respectively.

We apply our feature-extraction pipeline, followed by the proposed motion estimation algorithm, to one or more simulated images, and compare the estimated velocities with the ground-truth simulation parameters. For translational velocity, we perform a coarse grid search over initial hypotheses \{-25, 0, 25\} m/s along each axis. The spin optimization is initialized at $\boldsymbol{\omega}_{co} = [0,0,0]$.

We evaluate four scenarios: (1) a single image; (2) two images captured simultaneously by a stereo pair with a 10\,cm baseline; (3) two images captured by the same camera with a 2\,ms inter-frame delay;  and (4) two images captured by the same stereo pair but with 2\,ms inter-frame delay. All scenarios are evaluated on the same set of 100 simulation trials. We express all quantities in the camera coordinate frame, where the $x$-axis corresponds to the horizontal image direction, the $y$-axis to the vertical image direction, and the $z$-axis to the camera viewing direction.

Since existing rolling-shutter methods do not estimate 3D translational and angular velocities without explicit 3D--2D correspondences, we compare our approach with the nonlinear formulation of Ait-Aider et al. \cite{AitAider2006RSPoseVelocity}, which provides a solution for general objects. We provide the baseline method with ground-truth 3D--2D correspondences, while our method operates without correspondence information. To ensure comparable computational budget, both methods are initialized using the same coarse grid search strategy over motion hypotheses. In addition, we include results from directly applying a global-shutter perspective-n-point (PnP) method as a reference baseline.

\subsection{Real-World Setup and Baseline}
We evaluate our method in a challenging real-world application: a vision-based golf launch monitor. A well-struck golf ball can reach speeds of up to 70\,m/s and spin rates of up to 10{,}000\,rpm. Our overall device design is illustrated in \cref{fig:hardware}. We build the prototype using two 6.5 MP rolling-shutter cameras with a fixed baseline. The cameras are oriented with a 90-degree in-plane rotation so that the scan direction opposes the dominant ball motion, relaxing trigger-timing constraints. Otherwise, the ball may leave the field of view before the frame readout is completed. Since the dominant velocity direction is approximately known in this setup, we directly initialize the optimization, omitting the coarse grid search used in simulation.

We evaluate accuracy by placing our device alongside a commercially available launch monitor (SkyTrak~\cite{SkyTrak}) and comparing estimated velocity over the same set of shots. Note that SkyTrak is used as an external baseline rather than ground truth. According to its published specifications, the stated measurement tolerances are 
$\pm 1\,\mathrm{mph}$ for ball speed, 
$\pm 1^\circ$ for launch angle, 
$\pm 2^\circ$ for side angle, 
and $\pm 250\,\mathrm{rpm}$ for both  backspin and sidespin. 

\section{Results}
In this section, we first evaluate the proposed method on synthetic data, followed by experiments on real-world measurements where we compare our results with an external baseline.

\subsection{Synthetic Evaluation} \label{sec:syn_eval}

  \begin{table}[t] 
  \centering
  \renewcommand{\arraystretch}{1.15}
  \setlength{\tabcolsep}{2pt}
  \small
  \resizebox{\linewidth}{!}{%
  \begin{tabular}{l ccc ccc ccc ccc}
  \toprule
  & \multicolumn{3}{c}{1. Single}
  & \multicolumn{3}{c}{2. Stereo}
  & \multicolumn{3}{c}{3. Multi-Frame}
  & \multicolumn{3}{c}{4. Stereo MF} \\
  \cmidrule(lr){2-4}
  \cmidrule(lr){5-7}
  \cmidrule(lr){8-10}
  \cmidrule(lr){11-13}
  & $x$ & $y$ & $z$
  & $x$ & $y$ & $z$
  & $x$ & $y$ & $z$
  & $x$ & $y$ & $z$ \\
  \midrule
  Global Shutter
    & - & - & -
    & - & - & -
    & 11.57 & 8.03 & 30.58
    & 38.42 & 8.55 & 39.21 \\ 
  Ait-Aider\etal\cite{AitAider2006RSPoseVelocity}
    & 3.24 & 10.14 & 14.33
    & 3.25 & 7.80 & 10.87
    & 1.44 & 1.12 & 12.40
    & 2.76 & 1.16 & 9.84 \\ 

  Ours
    & 0.23 & 4.61 & 2.95
    & 0.12 & 0.23 & 0.19
    & 0.16 & 0.12 & 0.36
    & 0.19 & 0.10 & 0.26 \\ 

  \bottomrule
  \end{tabular}
  }
  \vspace{0.3\baselineskip}
  \caption{MAE of the translational velocity components for Scenarios 1--4 (m/s).}
  \label{tab:velocity}
  \end{table}

  \begin{table}[t] 
  \centering
  \renewcommand{\arraystretch}{1.15}
  \setlength{\tabcolsep}{2pt}
  \resizebox{\linewidth}{!}{%
  \begin{tabular}{l ccc ccc ccc ccc}
  \toprule
  & \multicolumn{3}{c}{1. Single} & \multicolumn{3}{c}{2. Stereo} & \multicolumn{3}{c}{3. Multi-Frames} & \multicolumn{3}{c}{4. Stereo MF} \\
  \cmidrule(lr){2-4}\cmidrule(lr){5-7}\cmidrule(lr){8-10}\cmidrule(lr){11-13}
  & $x$ & $y$ & $z$ & $x$ & $y$ & $z$ & $x$ & $y$ & $z$ & $x$ & $y$ & $z$ \\
  \midrule
   Global Shutter   
    & - & - & - & - & - & - & 131.00 & 126.50 & 120.88 & 132.52 & 123.50 & 119.31 \\
   Ait-Aider\etal\cite{AitAider2006RSPoseVelocity}   
    & 164.16 & 172.60 & 138.25 & 163.96 & 188.64 & 137.85 & 142.31 & 133.22 & 81.99 & 153.16 & 151.77 & 114.63 \\
   Ours 
    & 49.82 & 17.18 & 27.89 & 33.11 & 20.60 & 10.13 & 8.48 & 7.22 & 7.51 & 13.58 & 23.98 & 6.68 \\
  Ours + Ref.   
    & - & - & - & - & - & - & 4.58 & 2.41 & 1.10 & 10.39 & 3.93 & 2.68 \\
  \bottomrule
  \end{tabular}
  }
  \vspace{0.3\baselineskip}
  \caption{MAE of the angular velocity components for Scenarios 1--4 (rad/s).}
  \label{tab:spin}
  \end{table}

  \begin{figure}[t]
  \centering
  \begin{subfigure}{0.66\linewidth}
    \centering
    \includegraphics[width=\linewidth]{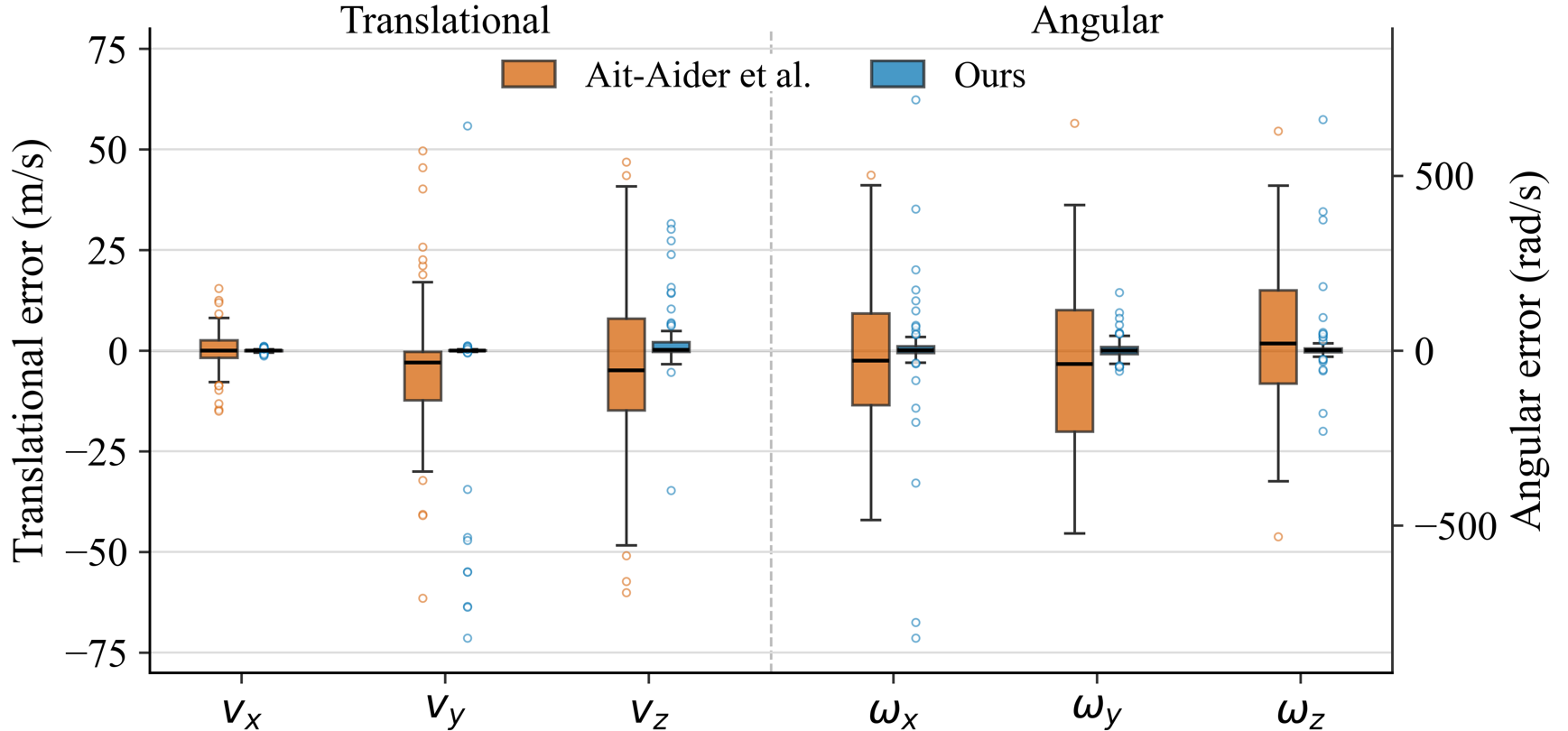}
    \caption{Error distribution with single frame.}
    \label{fig:distribution}
  \end{subfigure}
  \begin{subfigure}{0.33\linewidth}
      \centering
      \includegraphics[width=\linewidth]{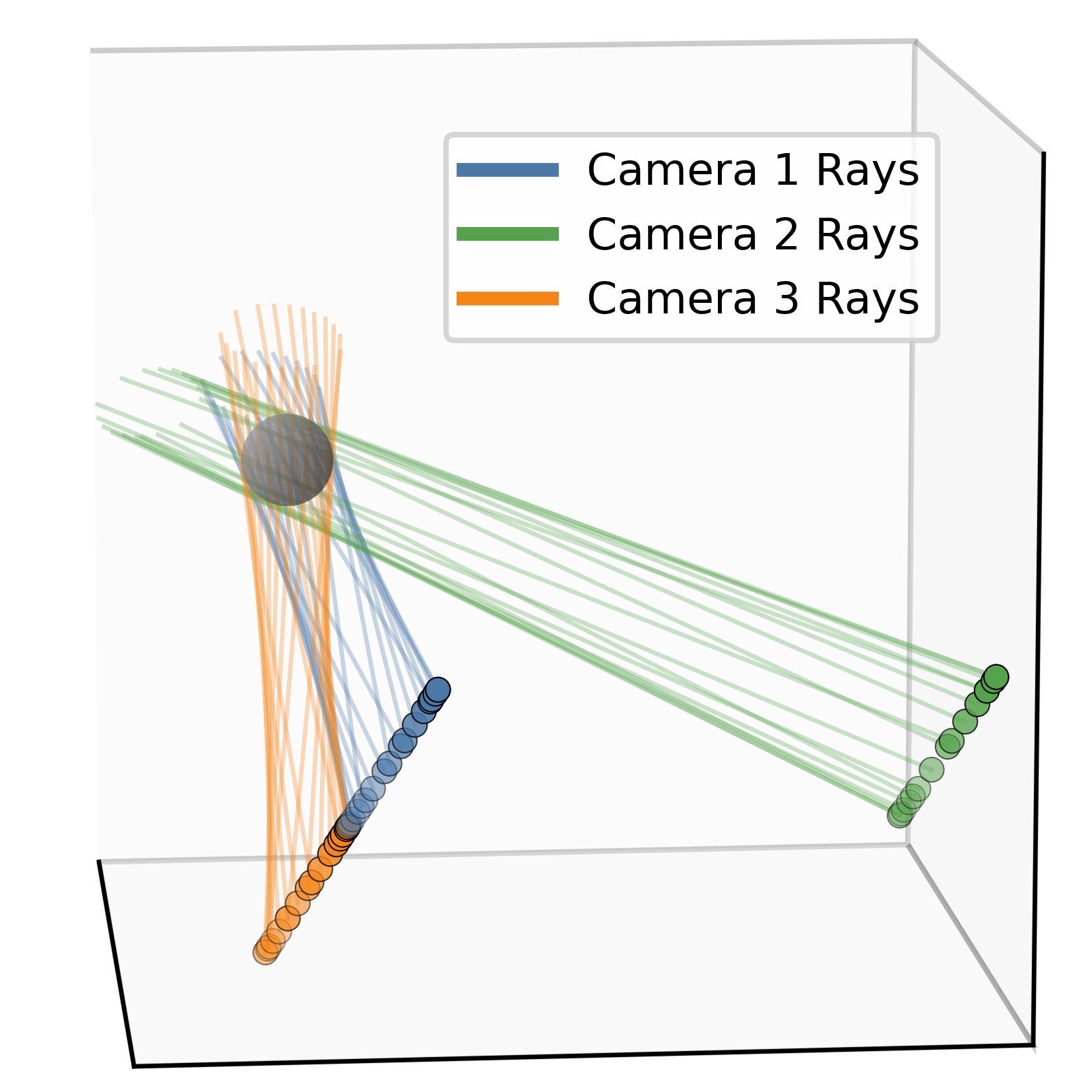}
      \caption{Multi-view constraints.}
      \label{fig:multi_velocity}
  \end{subfigure}
  \caption{(a) Boxplots of the error distribution obtained with the single-frame, single-camera setup. We report the mean and standard deviation of velocity errors for both the baseline and our method. (b) Multi-view constraints for velocity estimation. Motion primarily along the camera depth axis is ambiguous in a single rolling-shutter image (Camera~1). A second frame (Camera~2) or an additional viewpoint (Camera~3) provides complementary constraints that help resolve this ambiguity.}
  \end{figure}

  Following our two-step optimization strategy, we first estimate translational velocity and report the mean absolute errors (MAE) in \cref{tab:velocity}. We then fix the estimated velocities and optimize for angular velocity. The MAE results for angular velocity are reported in \cref{tab:spin}. The results demonstrate that, across all configurations, our proposed system consistently outperforms both Ait-Aider et al.~\cite{AitAider2006RSPoseVelocity} and the global-shutter baselines by a substantial margin in both translational and angular velocity estimation. Under the same rolling-shutter setup, the method of Ait-Aider et al.~\cite{AitAider2006RSPoseVelocity} exhibits large errors even when provided with ground-truth correspondences, due to its sensitivity to initialization under high translational and angular velocities.

  We also observe that our algorithm exhibits larger mean errors and variance along the $y$ and $z$ axes in the single-camera setting, as illustrated in \cref{fig:distribution}. This behavior is expected because motion along the $y$ and $z$ axes can produce similar rolling-shutter distortions, leading to ambiguities when observed from a single viewpoint. To resolve this ambiguity, introducing a second view or an extra frame provides additional constraints, as shown in \cref{fig:multi_velocity}, resulting in more reliable velocity estimates. The improvement from the two-camera configuration is more noticeable for rotational velocity, as illustrated in \cref{tab:spin}, which benefits from the additional spin-refinement step (\cref{fig:multi_spin}).

  \begin{table}[b]
  \centering
  \setlength{\tabcolsep}{8pt}
  {\renewcommand{\arraystretch}{0.9}
  \small
  \begin{tabular}{l ccc ccc}
  \toprule
  & $\mathbf{e}_{\mathbf{v}_x}$ & $\mathbf{e}_{\mathbf{v}_y}$ & $\mathbf{e}_{\mathbf{v}_z}$
  & $\mathbf{e}_{\boldsymbol{\omega}_x}$ & $\mathbf{e}_{\boldsymbol{\omega}_y}$ & $\mathbf{e}_{\boldsymbol{\omega}_z}$ \\
  \midrule
  80\% radius
    & 0.32 & 3.46 & 4.90
    & 45.32 & 31.54 & 44.02\\
  100\% radius
    & 0.23 & 4.61 & 2.95
    & 49.82 & 17.18 & 27.89\\
  120\% radius
    & 0.19 & 3.50 & 3.63
    & 38.14 & 14.86 & 22.28\\
  \bottomrule
  \end{tabular}%
  }
  \vspace{0.3\baselineskip}
  \caption{Effect of radius on $\mathbf{v} (m/s)$ and $\boldsymbol{\omega} (rad/s)$ error in single-image setting.}
  \label{tab:radius}
  \end{table}

  Additionally, we conduct an ablation study on the robustness of our approach with respect to different sphere radii. We evaluate settings with varying radii (80\%, 100\%, and 120\% of the default radius) in the single-image setting. As shown in \cref{tab:radius}, a larger radius generally yields smaller errors due to higher image resolution of the ball. However, for the largest radii, slight increases in the errors of $\mathbf{v}_y$, $\mathbf{v}_z$, and $\boldsymbol{\omega}_x$ are observed due to increased variance (\cref{fig:distribution}).

\subsection{Real-World Application Evaluation}
  For the real-world experiments, we recorded 10 different golf shots using both our system and a commercially available launch monitor (SkyTrak~\cite{SkyTrak}). The estimated translational velocity is converted to the speed--launch/side--angle representation reported by SkyTrak. The third spin component is omitted, as it is not provided by the device.
  
  We present the full comparison in \cref{tab:real_eval}. Across all 10 shots, our method produces speed and launch/side-angle estimates that closely match those reported by SkyTrak, demonstrating strong overall consistency across real-world trials. Spin estimates also show good agreement for 9 out of 10 shots, indicating reliable recovery of rotational dynamics under most conditions. However, shot~10 exhibits a noticeable discrepancy in spin, suggesting a potential inconsistency between the two systems.

\begin{figure}[t]
  \centering
  \begin{minipage}{0.42\linewidth}
    \centering
    \includegraphics[width=\linewidth]{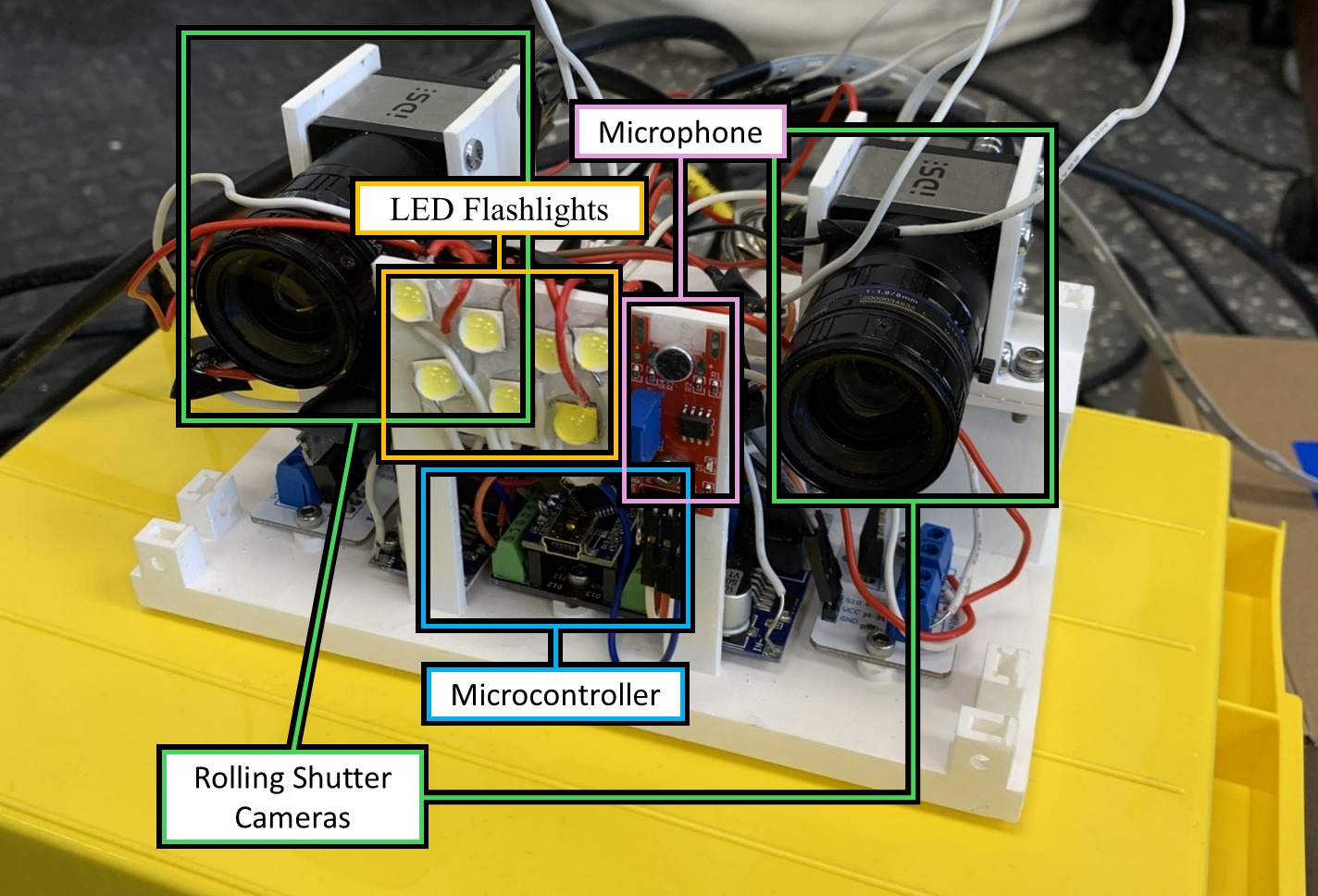}
    \subcaption{}
    \label{fig:hardware}
  \end{minipage}\hfill
  \begin{minipage}{0.57\linewidth}
    \centering
    \includegraphics[width=\linewidth]{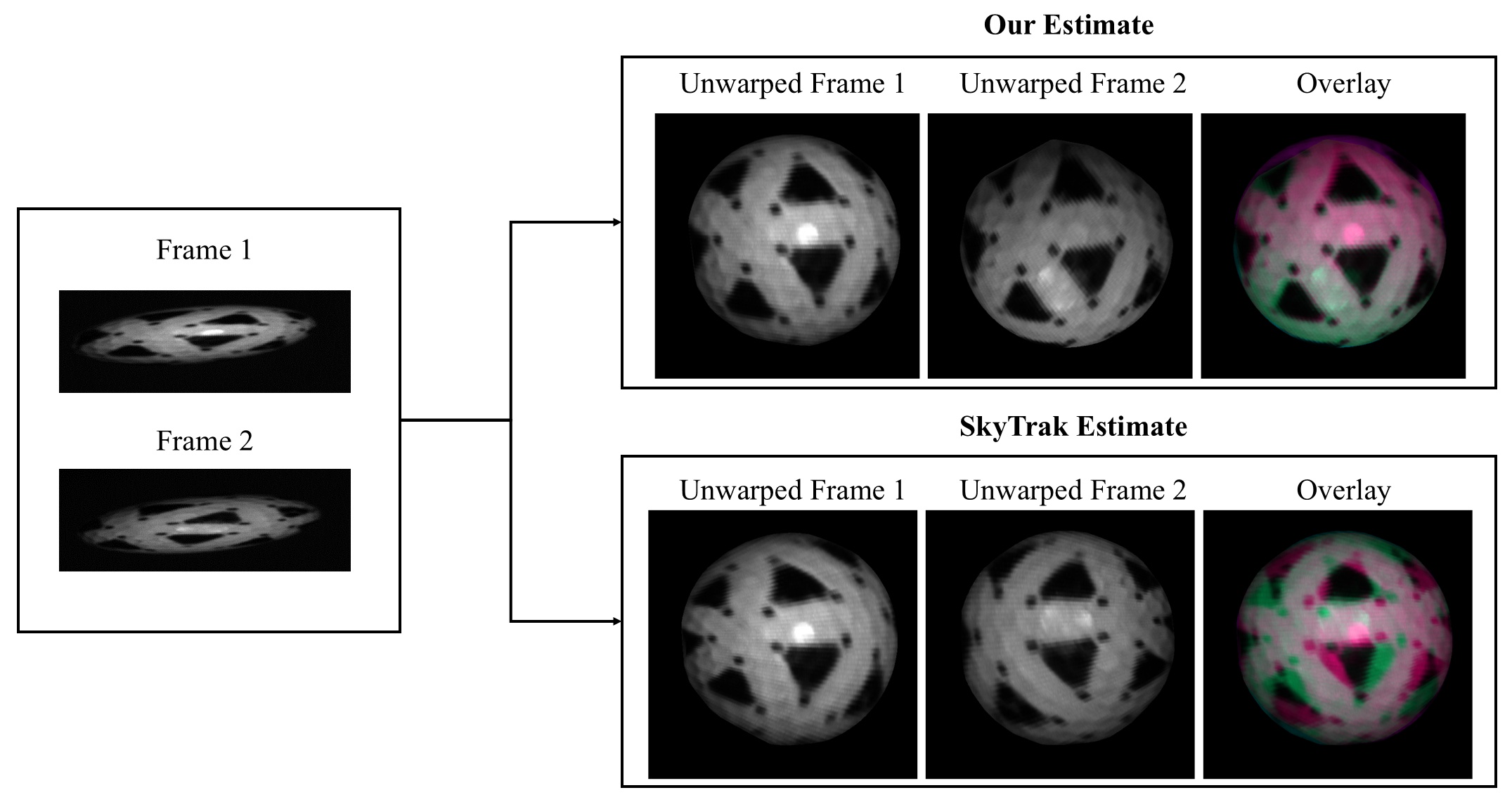}
    \subcaption{ }
    \label{fig:comparison}
  \end{minipage}
  \vspace{-0.5\baselineskip}
  \caption{
    Our prototype setup and qualitative comparison for an outlier case. (a) Our prototype hardware setup. The flash and cameras are triggered by the impact sound recorded by a microphone. The two rolling-shutter cameras are vertically mounted and captured with a 2\,ms inter-frame delay. (b) We validate the estimated spin by overlaying the unwarped images (red and green) after warping them using the estimated angular velocity. Accurate estimates result in strong alignment of the pattern.
  }
\end{figure}

  To further examine the abnormal result from shot~10, we unwarp the captured images using the estimated velocities and qualitatively inspect the resulting reconstructions. As shown in \cref{fig:comparison}, the reconstruction produced by our method exhibits stronger temporal consistency. Since SkyTrak estimates spin from consecutive global-shutter frames, differences in texture alignment may contribute to the observed discrepancy. In contrast, our approach leverages rolling-shutter distortions as additional constraints, which can help resolve such ambiguities.

\begin{table*}[t]
\centering
\renewcommand{\arraystretch}{1.15}

\resizebox{\textwidth}{!}{%
\begin{tabular}{c|ccc|ccc|ccc|ccc|ccc}
\toprule
\multirow{2}{*}{Data}
& \multicolumn{3}{c|}{Speed (mph)}
& \multicolumn{3}{c|}{Launch Angle (deg)}
& \multicolumn{3}{c|}{Side Angle (deg)}
& \multicolumn{3}{c|}{Backspin (rpm)}
& \multicolumn{3}{c}{Sidespin (rpm)}\\
\cmidrule(lr){2-4}\cmidrule(lr){5-7}\cmidrule(lr){8-10}\cmidrule(lr){11-13}\cmidrule(lr){14-16}
& Ours & SkyTrak & Diff
& Ours & SkyTrak & Diff
& Ours & SkyTrak & Diff
& Ours & SkyTrak & Diff
& Ours & SkyTrak & Diff\\
\midrule
1  & 107.4 & 106  & 1.4  & 25.6 & 25 & 0.6 & 0.9  & -3 & 3.9  & 3856 & 3583 & 273   & 698 & 780  & -82\\
2  & 97.2  & 96   & 1.2  & 25.6 & 25 & 0.6 & -1.3 & -6 & 4.7  & 3731 & 3794 & -63   & 530 & 0    & 530\\
3  & 122.2 & 120  & 2.2  & 26.2 & 25 & 1.2 & -1.4 & 5  & -6.4 & 4320 & 4150 & 170   & 732 & 237  & 495\\
4  & 113.8 & 107  & 6.8  & 27.3 & 27 & 0.3 & 2.8  & 3  & -0.2 & 4255 & 4159 & 96    & 482 & 0    & 482\\
5  & 118.2 & 116  & 2.2  & 25.6 & 24 & 1.6 & 3.7  & 0  & 3.7  & 4408 & 4511 & -103  & 495 & 453  & 42\\
6  & 55.1  & 54   & 1.1  & 26.6 & 26 & 0.6 & 2.9  & 0  & 2.9  & 2741 & 2664 & 77    & 602 & 383  & 219\\
7  & 99.3  & 95   & 4.3  & 24.2 & 23 & 1.2 & 0.6  & -3 & 3.6  & 4365 & 4249 & 116   & 858 & 798  & 60\\
8  & 69.7  & 70   & -0.3 & 27.8 & 26 & 1.8 & -7.4 & -4 & -3.4 & 3269 & 3208 & 61    & 468 & 368  & 100\\
9  & 100   & 98   & 2    & 25.3 & 24 & 1.3 & 1.5  & 4  & -2.5 & 4215 & 4218 & -3    & 518 & 302  & 216\\
10 & 86.3  & 89   & -2.7 & 25.5 & 25 & 0.5 & 0.2  & 0  & 0.2  & 3805 & \textcolor{red}{11254} & \textcolor{red}{-7449} & 667 & \textcolor{red}{-1129} & \textcolor{red}{1796}\\
\midrule
\multicolumn{1}{c|}{MAE}
& \multicolumn{2}{c}{} & 2.4
& \multicolumn{2}{c}{} & 1.0
& \multicolumn{2}{c}{} & 3.2
& \multicolumn{2}{c}{} & 841
& \multicolumn{2}{c}{} & 402 \\
\multicolumn{1}{c|}{MAE w/o \#10}
& \multicolumn{2}{c}{} & 2.4
& \multicolumn{2}{c}{} & 1.0
& \multicolumn{2}{c}{} & 3.5
& \multicolumn{2}{c}{} & 107
& \multicolumn{2}{c}{} & 247 \\
\bottomrule
\end{tabular}
}
\vspace{0.3\baselineskip}
\caption{Comparison between our device and SkyTrak~\cite{SkyTrak} on 10 golf shots. Differences are computed as (Ours -- SkyTrak). Data~10 is treated as an outlier, and its anomalous measurements are highlighted in red.}
\label{tab:real_eval}
\end{table*}

\section{Conclusion}

In this work, we present a correspondence-free formulation for estimating the 3D translational and angular velocity of spherical objects from rolling-shutter imagery. Rather than correcting rolling-shutter distortions, we exploit them as a rich source of temporal information for velocity estimation. We design a robust surface pattern on the sphere and formulate the estimation as a backward-projection optimization. This avoids explicit 3D--2D correspondences on repetitive spherical textures while still exploiting the structured distortions induced by rolling-shutter readout. We adopt a two-stage optimization strategy that explicitly decouples translational and rotational estimation, improving robustness while reducing optimization complexity. The formulation extends naturally from single-image to multi-image and multi-camera settings, where increased temporal baselines and additional viewpoints provide stronger geometric constraints. Experiments in both simulation and real-world scenarios demonstrate accurate motion recovery across a wide range of configurations.

There are several limitations of our proposed approach. First, although it does not require explicit correspondences, the system still depends on additional known parameters, such as the sphere radius, and on a carefully designed surface pattern. Second, our method is currently restricted to spherical objects, rather than other non-spherical geometries. Third, we assume high-speed motion and model the sphere’s motion with a constant-velocity assumption, which may limit its applicability to more complex, time-varying motion.

In future work, we will relax the assumption of known parameters. Moreover, we will extend our method beyond spherical objects to more general rigid bodies, including ellipsoids and other non-spherical shapes.

\FloatBarrier

%
%
\bibliographystyle{splncs04}
\bibliography{main}
\end{document}